# Promotion of Answer Value Measurement With Domain Effects in Community Question Answering Systems

Binbin Jin, Enhong Chen, *Senior Member, IEEE*, Hongke Zhao, Zhenya Huang, Qi Liu, *Member, IEEE*, Hengshu Zhu, *Member, IEEE*, and Shui Yu

*Abstract*—In the area of community question answering (CQA), answer selection and answer ranking are two tasks which are applied to help users quickly access valuable answers. Existing solutions mainly exploit the syntactic or semantic correlation between a question and its related answers (Q&A), where the multifacet domain effects in CQA are still underexplored. In this paper, we propose a unified model, enhanced attentive recurrent neural network (*EARNN*), for both answer selection and answer ranking tasks by taking full advantages of both Q&A semantics and multifacet domain effects (i.e., topic effects and timeliness). Specifically, we develop a serialized long short-term memory to learn the unified representations of Q&A, where two attention mechanisms at either sentence level or word level are designed for capturing the deep effects of topics. Meanwhile, the emphasis of Q&A can be automatically distinguished. Furthermore, we design a time-sensitive ranking function to model the timeliness in CQA. To effectively train *EARNN*, a question-dependent pairwise learning strategy is also developed. Finally, we conduct extensive experiments on a real-world dataset from Quora. Experimental results validate the effectiveness and interpretability of our proposed *EARNN* model.

## I. INTRODUCTION

**W**ITH the prevalence of community question answering (CQA), e.g., Quora[1] and Yahoo Answers,[2] there

Manuscript received September 6, 2018; revised January 24, 2019; accepted May 4, 2019. This work was supported in part by the National Key Research and Development Program of China under Grant 2016YFB1000904, in part by the National Natural Science Foundation of China under Grant U1605251, Grant 61727809, and Grant 61672483, and in part by the Youth Innovation Promotion Association of Chinese Academy of Sciences under Grant 2014299. This paper was recommended by Associate Editor Y. Zhao. *(Corresponding author: Enhong Chen.)*

B. Jin, E. Chen, Z. Huang, and Q. Liu are with the School of Computer Science and Technology, University of Science and Technology of China, Hefei 230026, China (e-mail: bb0725@mail.ustc.edu.cn; cheneh@ustc.edu.cn; huangzhy@mail.ustc.edu.cn; qiliuql@ustc.edu.cn).

H. Zhao is with the College of Management and Economics, Tianjin University, Tianjin 300072, China (e-mail: hongke@tju.edu.cn).

H. Zhu is with the Baidu Talent Intelligence Center, Baidu, Inc., Beijing 100085, China (e-mail: zhuhengshu@baidu.com).

S. Yu is with the School of Computer Science, University of Technology Sydney, Sydney, NSW 2007, Australia (e-mail: shui.yu@uts.edu.au).

Color versions of one or more of the figures in this paper are available online at http://ieeexplore.ieee.org.

Digital Object Identifier 10.1109/TSMC.2019.2917673

[1]https://www.quora.com/
[2]https://answers.yahoo.com/

are more and more users active on these communities. They are willing to post questions or share their experience so that massive questions with largely increasing answers are accumulated. In CQA websites, after posting questions, many answerers will continuously contribute to their interested ones. Particularly, some attractive questions can appeal to even hundreds of answers, and also, some of them may make users lose their interest due to their long paragraphs [1]. Thus, it is hard for the askers and attracted visitors to quickly find the valuable answers, especially for those answers whose views and upvotes are not stable yet. Therefore, answer selection and answer ranking, which can be applied to similar application scenarios, have become two effective solutions. Particularly, both of them focus on measuring the semantic matching between the question and its related answers (Q&A). The difference is that answer selection aims to select valuable answers from the candidate list, while answer ranking aims to put valuable answers ahead in the answer list [2]. However, as hot research points in the area of CQA, answer selection and ranking are still with challenges.

These challenges mainly come from two aspects. First, for the past few years, researchers have proposed some deep learning-based models with attention mechanisms to exploit the semantics between Q&A [3]–[5]. These studies mainly use the text of Q&A to measure their similarity. Recent years, with the mechanism improvements of CQA sites, almost all platforms provide topics for each question (see the blue rectangle in Fig. 1) so that visitors can quickly find their interested ones with topic filtering. However, exploiting the benefits of these topics for answer selection or ranking is still open and being explored.

Second, some studies mainly focus on fixing the lexical gap between Q&A on large-scale dataset, such as Yahoo Answers or Quora [6]–[8]. However, they ignore the fact that different answers are posted at different times so that their value is highly related to the time [9]. Therefore, it is necessary to eliminate the bias induced by the time.

In order to solve the challenges, we make a deep observation and exploration in CQA and then find multifacet domain effects. Fig. 1[3] explains the significance of two domain effects related to our motivations with a snapshot from Quora. The

[3]This question and corresponding answers are posted on the following url: https://www.quora.com/Which-are-some-of-the-best-places-to-visit.

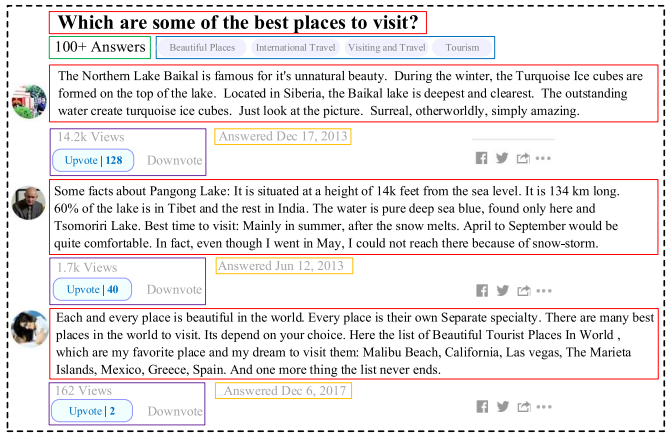

Fig. 1. Example of Q&A from Quora with a question and three comparable answers.

figure shows a popular question which indeed receives more than 100 answers (green rectangle) and we only show three compared answers for better illustration. From this instance, besides "Details of Q&A (red)," we can also see "Topics (blue)," "Answer Time (orange)," and "Views and Upvotes (purple)" in rectangles. Obviously, three sample answers have different value[4] (14.2k views and 128 upvotes *versus* 1.7k views and 40 upvotes *versus* 162 views and 2 upvotes). According to our observations, there are at least two aspects which have great impacts on measuring answer value.

*Topic Effects:* The value of an answer not only depends on its description but also is affected by the topics. Different from some platforms (e.g., Yahoo Answers) which only provide predefined topics, Quora allows askers to create any topic for their questions. As a consequence, askers will use more exact topics that are relevant to their questions. It is obvious that there will be several key words in the topics and Q&A, and these key words, or the emphasis, can help visitors quickly understand the implicit intent of Q&A [10]. For example, we can see that the first and second answers in Fig. 1 refer to some places which are the intent of the asker, whereas only the first one receives more upvotes. To explain this difference, we find the first good answer mentions more words which are similar with "beautiful" appeared in topics, e.g., "beauty," "surreal," etc. Especially the last sentence ("Surreal, otherworldly, simply amazing.") is likely to attract more people. On the contrary, even though the second one involves "Pangong Lake," the description is bland and does not appeal to readers. Thus, with the help of topics, we can better understand the deep semantic and distinguish the emphasis of Q&A.

*Timeliness of Answers:* As we mentioned, in CQA, the number of upvotes depends on not only the quality of answers but also the time period since answers are posted. We call the latter fact timeliness. Intuitively, a question will attract more readers during the early period after it is posted. Consequently, the corresponding early coming answers are supposed to serve more readers and receive more upvotes. In Fig. 1, the first and the third answers are both good ones with high quality, but the timestamp of the first answer is much earlier than the third one's so that the former is viewed and voted with more chances than the latter. Therefore, when measuring answer value, it is significant to take the timeliness into account.

Considering both topic effects and timeliness of answers, it is necessary to find an approach which can well model these special domain effects on understanding the deep semantics of Q&A and evaluating the answer value. Also, to conveniently access Q&A, especially for those long answers, it is significant to distinguish the emphasis of Q&A with the help of topics.

To that end, in this paper, we present a focused study on answer selection and ranking by taking full advantages of both Q&A and two specific domain effects (i.e., topic effects and timeliness). Specifically, we propose a unified model, enhanced attentive recurrent neural network (*EARNN*), to exploit the impacts of topics and timeliness on evaluating answer value. Particularly, we first follow the question answering process and develop a serialized long short-term memory (LSTM) with two enhanced attention mechanisms to capture the deep effects of topics. Benefiting from our attention mechanisms, we can easily find the important regions which are related to the intent of the asker (i.e., topics). After that the unified representations of Q&A are learned and the emphasis of Q&A can be automatically distinguished at sentence and word levels. Moreover, considering the timeliness that answers with earlier timestamps are supposed to be more preferred, we develop a time-sensitive ranking function to eliminate the bias induced by the time (i.e., timeliness). Furthermore, since visitors of different questions may vary a lot, it is usually unreasonable to compare the value of answers to different questions. Thus, we adopt a question-dependent pairwise learning strategy to facilitate the training process of our model. Finally, we construct extensive experiments on a real-world dataset. The experimental results validate the effectiveness and interpretability of *EARNN*. The contributions of this paper can be summarized as follows.

1)  We conduct a focused study on answer selection and ranking problems in CQA. We further propose a unified model (i.e., *EARNN*) for both tasks to effectively measure answer value.
2)  We make a deep observation and find the topic effects which have an impact on measuring semantic relations between Q&A. Therefore, we propose two enhanced attention mechanisms to capture the deep effects of topics. Benefiting from them, the emphasis of Q&A can be automatically distinguished at sentence and word levels.
3)  To eliminate the bias induced by the time, we develop a time-sensitive rank function to model timeliness of answers.
4)  We collect large-scale real-world data from Quora. With this data, we conduct extensive experiments whose results demonstrate the effectiveness of *EARNN*.

## II. Related Work

In CQA, the related work can be grouped into two categories. One is the traditional methods which mainly depend on lots of manual work. Another is neural network-based methods

---

[4]The value means the attraction to readers, which is often reflected in the numbers of their stable views and upvotes.

which avoid feature engineering and their performances have been validated.

### A. Feature-Based Approaches

In the early stage, researchers tried their best to design various nontextual features to predict answer quality, including answer length, answerer's activity level, question–answer overlap, and so on [9], [11]–[15]. Then, with the development of natural language processing (NLP) [16], [17], many lexical and syntactic-based approaches were applied to analyze the structure of sentences and the relations between Q&A. Yih *et al.* [18] paid attention to improving the lexical semantics based on word relations, including synonymy/antonymy, hypernymy/hyponymy, and general semantic word similarity. For syntactic approaches, dependency trees [19]–[21] or quasi-synchronous grammar [22], [23] was used to analyze the structure of sentences and extract effective syntactic features. Besides, Cai *et al.* [24] and Ji *et al.* [25] adopted the topic models to extract topic distributions as contextual features under the assumption that the question and answer should share a similar topic distribution. After the feature engineer, logistic regression (LR) [14], support vector machine (SVM) [21], conditional random fields (CRFs) [26], [27], and other machine learning methods [28] were employed to measure answer quality. Since Jeon *et al.* [29] proposed a word-based translation model for question retrieval, researchers begun to make their efforts on fixing the lexical gap between Q&A. Xue *et al.* [30] proposed a word-based translation language model for question retrieval and Lee *et al.* [31] tried to improve the translation probabilities based on question–answer pairs by selecting the most important terms to build compact translation models. Furthermore, phrase-based models [32], [33] and lexical word-based translation models [34] were proposed in succession for better measuring the similarities of Q&A. In summary, those feature engineering methods depend on much manual work which is time consuming. The translation-based methods suffer from the informal words or phrases in Q&A archives and perform less applicability in new domains.

### B. Neural Network-Based Approaches

Recently, with the development of deep learning, scholars proposed some neural network-based models to explore the semantic relations of Q&A texts [35], [36], which achieved great success on measuring answer quality [3], [4], [37]. Despite the deep belief networks (DBNs) [38] and recursive neural networks [39] have shown some nonlinear fitting capability, the great success of convolutional neural networks (CNNs) [40]–[42] and recurrent neural networks (RNNs) [43]–[45] on various tasks completely changed the research direction. For example, Severyn and Moschitti [3] employed a CNN to generate a representation for each sentence and then used similarity matrix to compute a relevant score. Wang and Nyberg [36] proposed a method using a stacked bidirectional LSTM to read the sentence word-by-word, and then integrated all outputs to predict the quality score. Wan *et al.* [46] combined the LSTM and CNN to capture both local and context information for determining the importance of local keywords from the whole sentence view.

### TABLE I
### SEVERAL IMPORTANT MATHEMATICAL NOTATIONS

| Notations | Type | Description |
|---|---|---|
| $N$ | scalar | the number of words in a sentence |
| $M$ | scalar | the number of sentences in an answer |
| $C$ | vector | $C_i$ is the number of words in $i$-th topic |
| $s$ | matrix | the representation of a sentence in Q&A |
| $c$ | tensor | the representation of topics |
| $q^r$ | matrix | the outputs of LSTM_Q |
| $a^r$ | tensor | the outputs of LSTM_A |
| $q^f$ | vector | the unified representation of the question |
| $a^f$ | vector | the unified representation of the answer |
| $c^f$ | vector | the unified representation of topics |
| $\hat{V}$ | scalar | the matching score of the answer |
| $t$ | scalar | the time period since the answer is posted |
| $\widetilde{V}$ | scalar | the ranking score of the answer |

In a word, CNN-based models use position-shared weights with local perspective filters to learn spatial regularities in Q&A while RNN-based models pay more attention on the regularities of the word sequence.

Moreover, in order to deeply exploit the semantic relevance between Q&A, some researchers attempted to integrate attention mechanisms into CNN/RNN-based models [5], [47]–[50]. These adjusted the machine's attention on different regions of texts so that the relations of Q&A could be better understood. For instance, Yin *et al.* [47] described three architectures where attention mechanisms were combined with a CNN for general sentence pair modeling tasks. Liu *et al.* [5] modeled strong interactions of two texts through two inter- and intra-dependent LSTMs. However, to the best of our knowledge, almost all neural network models only focus on modeling the similarities of Q&A and few works directly model multifacet domain effects, such as topic effects or timeliness.

Different from previous studies, this paper aims at modeling two specific domain effects in CQA, i.e., topic effects and timeliness, which are potentially beneficial for measuring answer value. Specifically, we develop a serialized LSTM with two enhanced attention mechanisms (i.e., sentence level and word level) to mine the deep effects of topics and further recognize the emphasis of Q&A. Besides, we design a time-sensitive ranking function to integrate timestamps into our proposed model *EARNN* and perceive the timeliness in CQA.

### III. METHODOLOGY

In this section, we first formally inspect answer selection and answer ranking problems. Then we introduce the technical details of our *EARNN*, including the architecture of the neural network and the training method. For better illustration, Table I lists some mathematical notations.

### A. Problem Overview

Generally, answer selection mainly targets at choosing valuable answers from candidates while answer ranking mainly targets at sorting answers by their value to a specific question. Specifically, given a question $q$, a list of answers $A = \{a^1, a^2, \ldots\}$ is followed, where $a^i$ is the $i$th answer. In addition, each question has several topics $c$ and each answer has a timestamp $t$ and ground truth $v$. Then, two tasks have the following formulations.

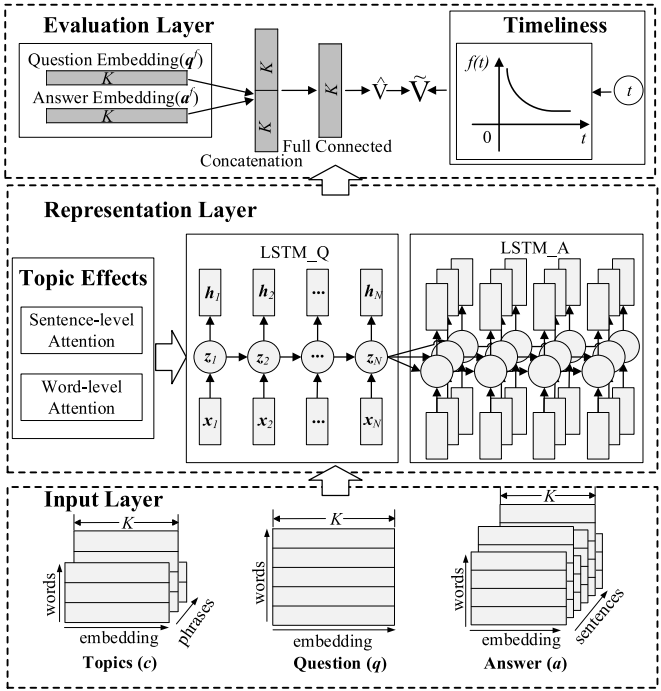

Fig. 2. Graphical representation of *EARNN*.

*Task 1 (Answer Selection):* Given a question $q$ with topics $c$ and a list of answers $A$ with a series of timestamps, our goal is to integrate all information to train a model (i.e., *EARNN*), which can be used to label all candidate answers with 0/1 (1 for valuable answer; 0 otherwise), s.t. the labels are consistent with the ground truth, such as manual annotation.

*Task 2 (Answer Ranking):* Given a question $q$ with topics $c$ and a list of unsorted answers $A$ with a series of timestamps, our goal is to integrate all information to train a model (i.e., *EARNN*), which can be used to rank all candidate answers, s.t. the order of sorted answers is consistent with their ground truth, such as the stable upvotes.

### B. Technical Details of EARNN

In this section, we introduce *EARNN* in detail. As shown in Fig. 2, *EARNN* contains three parts, namely, input layer, representation layer, and evaluation layer. Particularly, multifacet domain effects (i.e., topic effects and timeliness) are modeled in representation layer and evaluation layer, respectively. To effectively train *EARNN*, a question-dependent pairwise learning strategy is also proposed at the end of this section.

*Input layer* aims to represent words in Q&A in a continuous space. As mentioned in Section III-A, the textual inputs to *EARNN* include three parts (i.e., a question $q$, an answer $a$, and topics $c$). For simplicity, we assume the question $q$ contains only one intent of the asker so that it is treated as one sentence. Differently, answers usually settle a question from various perspectives, thus an answer $a$ is formalized as a sequence of $M$ sentences. For each sentence in Q&A, it is formalized as a sequence of $N$ words. Besides, topics $c$ consist of $|C|$ phrases and each phrase contains $C_j (j \in \{1, 2, \ldots, |C|\})$ words. Then, we replace each word with a pretrained $K$-dimensional word embedding [51]. After that as shown in Fig. 2, inputs $q$, $a$, and $c$ are composed of one or more matrices each of which represents a sentence or a phrase. Note that $N$, $M$, and $|C|$ are not fixed relaying on the instance and pretrained word embeddings only capture the syntactic information rather than the semantic one.

*Representation layer* develops a serialized LSTM with two enhanced attention mechanisms so that topic effects are modeled and unified representations of Q&A are learned. Meanwhile, the emphasis of Q&A is automatically distinguished.

In this paper, since the attraction of answers partly depends on the question, answer representations should be adjusted according to the question. Therefore, we develop a serialized LSTM built by two LSTM models (i.e., LSTM_Q and LSTM_A) as shown in Fig. 2. Specifically, given the question $q$ and answer $a$, LSTM_Q reads the word embeddings of the question one by one while LSTM_A (with different parameters from LSTM_Q) reads the word embeddings of the answer in a more sophisticated way. First, we initialize the memory cell of LSTM_A with the final memory cell of LSTM_Q to model the relations of Q&A. Then, since sentences in the answer may express different meanings, they should be independently modeled by LSTM_A. Therefore, each word in a sentence can learn a semantic word embedding by combining the question and its adjacent words.

With respect to the basic RNN cell, we utilize an implementation of LSTM proposed by Graves *et al.* [52]. Given the $t$-th word embedding $x_t$ ($t \in \{1, 2, \ldots, N\}$) and $(t-1)$th cell vector $z_{t-1}$ of a specific sentence, the hidden vector $h_t$ can be computed as

$$h_t = \text{LSTM}(h_{t-1}, z_{t-1}, x_t). \tag{1}$$

Then, all word representations of Q&A, denoted by $q^r$ and $a^r$, can be constructed by a set of hidden vector sequences $h_t$ generated from LSTM_Q and LSTM_A.

After learning the representations of Q&A, i.e., $q^r$ and $a^r$, in the following, we aim to learn the unified representations of Q&A combining with topic effects. In Fig. 1, the first good answer mentions "Baikal Lake" and uses beauty and surreal to describe a beautiful scenery, which are semantically relevant to words in the question (e.g., "place") and topics (e.g., beautiful). But the second answer only refers to Pangong Lake with its location and best visiting time. Its description is bland and boring so that it loses much attraction from askers and visitors even though the posting time of the second one is earlier than the first one. Based on this evidence, we first design a sentence-level attention to capture the relations of Q&A. Then, on the basis of it, we further design a word-level attention together with topics to locate the significant regions of Q&A which are semantically relevant to topics. Formally, given the question $q^r$, answer $a^r$, and topics $c$, with the help of these two attentions, the unified representations of Q&A, denoted by $q^f$ and $a^f$, will be learned and each word or sentence is assigned an attention score denoting its importance.

Specifically, for the question $q^r$, the sentence-level attention [Fig. 3(a)] applies an average pooling to summarize all $N$ words into a fixed-length vector which denotes the semantic summary of the question. Thus, the final question representation $q^f \in \mathbb{R}^K$ is implemented as follows:

$$q^f = \frac{1}{N} \sum_{i=1}^{N} q_i^r. \qquad (2)$$

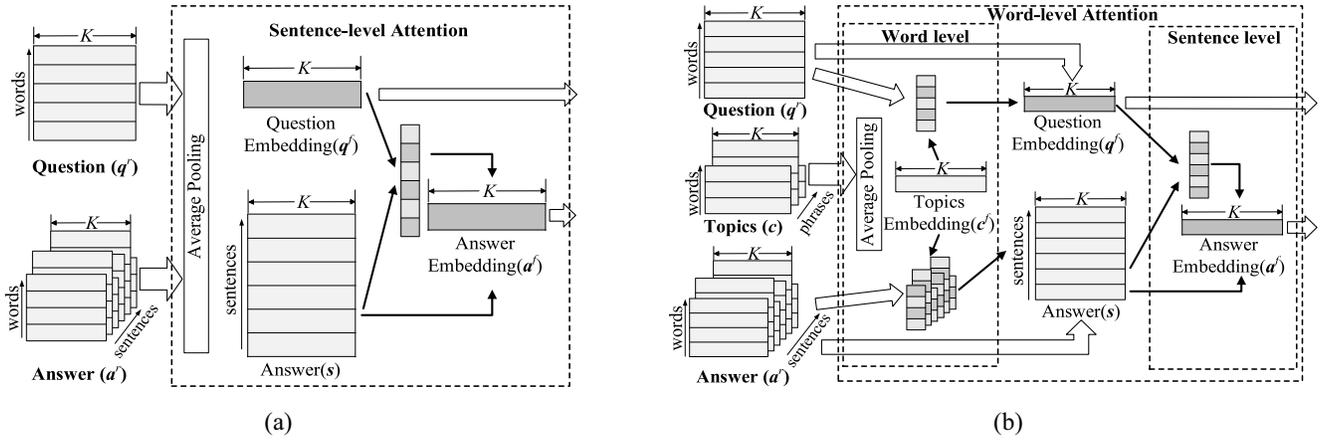

Fig. 3. Graphical representations of two attention mechanisms. (a) Sentence-level attention. (b) Word-level attention.

Similarly, for the answer $a^r$, the sentence-level attention first puts $j$th sentence $a_j^r$ into the average pooling and gets its semantic representation $s_j$ ($j \in \{1, 2, \ldots, M\}$). Then, a distance function is used to compute the attention score $\alpha_j$ for each sentence $s_j$ in the answer to the question $q^f$. Finally, the answer representation $a^f \in \mathbb{R}^K$ is modeled by a weighted sum of all $M$ sentences as follows:

$$a^f = \sum_{j=1}^{M} \alpha_j s_j, \quad s_j = \frac{1}{N} \sum_{i=1}^{N} a_{ji}^r, \quad \alpha_j = f\left(q^f, s_j\right) \qquad (3)$$

where $f(\cdot, \cdot)$ denotes the distance function which is stated as *cosine* similarity in this paper.

Differently, the word-level attention [Fig. 3(b)] can not only distinguish the important regions of the answer at sentence level but also focus on the word-level regions with the help of topics $c$. Given topics $c$ where $c_{ji}$ denotes the $i$th word in the $j$th phrase, the word-level attention applies an average pooling to summarize these topics into a fixed-length vector $c^f \in \mathbb{R}^K$

$$c^f = \frac{1}{|C|} \sum_{j=1}^{|C|} \left( \frac{1}{C_j} \sum_{i=1}^{C_j} c_{ji} \right). \qquad (4)$$

After the computation of the topic embedding $c^f$, word-level attention uses it to measure the attention score for each word in Q&A. Regarding the question $q^r$, since it contains a set of semantic representations whereas the topic embedding $c^f$ is a syntactic representation, a translation matrix $W$ is adopted to measure the distance between each word and topics. Then, a softmax operation follows to compute the attention score $\beta_i$ ($i \in \{1, 2, \ldots, N\}$) for $i$th word in the question. Finally, the final question representation $q^f \in \mathbb{R}^K$ can be modeled by a weighted sum as follows:

$$q^f = \sum_{i=1}^{N} \beta_i q_i^r, \quad \beta_i = \frac{\exp\left(h(c^f, q_i^r)\right)}{\sum_{k=1}^{N} \left(\exp\left(h(c^f, q_k^r)\right)\right)} \qquad (5)$$

where $h(a, b) = a^T W b$ denotes the distance between $a$ and $b$ in different spaces and the translation matrix $W$ is optimized by the network. As for the answer $a^r$, each sentence vector $s_j \in \mathbb{R}^K$ can be computed by the same way. Then we use (3) to get the final answer representation $a^f \in \mathbb{R}^K$.

In summary, the sentence-level attention utilizes the question and answer representations (i.e., $q^r$ and $a^r$) to distinguish important sentences of the answer. The word-level attention utilizes the extra topic representations (i.e., $c$) to capture the deep semantic relevance between topics and Q&A. Besides, according to the attention scores [i.e., $\alpha$ in (3) and $\beta$ in (5)], both important words and sentences can be recognized which will be demonstrated in experiments. After the process of this layer, the unified representations of Q&A (i.e., $q^f, a^f \in \mathbb{R}^K$) are obtained in a deep semantic space.

*Evaluation layer* outputs the answer value by combining the semantic matching of Q&A and the timeliness of answers. Actually, the timeliness means that the value of questions or answers will be reduced as time goes on. That is, a question will attract more readers in the early period after it is posted, so that the corresponding early coming answers may serve more readers and receive more upvotes. For example, in Fig. 1, the first and third answers both mention several beautiful places. They should have received the same attention. However, the first one receives more views and upvotes than the third one because it is posted much earlier than the other. According to this observation, we design a time-sensitive ranking function to model the biased value of answers. Specifically, given the question embedding $q^f$, answer embedding $a^f$, and timestamp $t$, we first measure the deep semantic matching score $\hat{V}$ of the answer and then take the timeliness into account to obtain the final ranking score $\widetilde{V}$.

For measuring the deep semantic matching score, we first concatenate the question $q^f$ and answer $a^f$ representations. Then, a fully connected network is used to learn the overall relevance representation $u$. Finally, a logistic function is applied to predict the deep semantic matching score $\hat{V}$

$$\hat{V} = \sigma(W_2 u + b_2), \quad u = \tau\left(W_1\left[q^f \oplus a^f\right] + b_1\right) \qquad (6)$$

where $\sigma(\cdot)$ and $\tau(\cdot)$ are sigmoid$(\cdot)$ and tanh$(\cdot)$ functions, respectively. $\oplus$ is the concatenation operation. $\{W_1, b_1, W_2, b_2\}$ are model parameters optimized by the network.

With respect to the timeliness of answers, we assume that an answer with an earlier timestamp $t$ is more valuable and attractive so that it is supposed to be ranked first than others. Therefore, we measure the revised ranking score $\widetilde{V}$ of the answer by jointly exploiting the relations between the deep semantic matching and timeliness

$$\widetilde{V} = \exp^{-\frac{t-t_0}{H}} \hat{V} \tag{7}$$

where $t_0$ is the timestamp of the first coming answer and $H$ is the hyperparameter. In (7), the first multiplier is a *decay factor* and it becomes smaller as time goes on. Particularly, it equals to 1 for the first answer.

*Model Training:* Since askers can post their questions at any time and visitors of different questions may vary a lot, it is usually unreasonable to compare the value of answers to different questions. In other words, we assume that only the value of answers to the same question is comparable. Thus, we adopt a question-dependent pairwise learning strategy with a large margin objective to optimize all parameters. First, for each question $q$, we construct several triples $(q, a^+, a^-)$ from the answer list, where answer $a^+$ is more valuable than answer $a^-$. Then, we minimize the following objective function:

$$\mathcal{L} = \min_{\Theta} \max(0, m + S(q, a^-) - S(q, a^+)) \tag{8}$$

where $\Theta$ is all parameters in *EARNN*; $S(q, a)$ denotes the ranking score $\widetilde{V}$ illustrated in (7); and $m$ is the margin which is a hyperparameter. Given a triple $(q, a^+, a^-)$, we will compute $\Delta S = m + S(q, a^-) - S(q, a^+)$. If $\Delta S \leq 0$, we will skip this triple. Otherwise, we use stochastic gradient decent (SGD) [53] to update the model parameters with the backpropagation through time algorithm.

## IV. EXPERIMENTS

In this section, we first introduce the dataset and show some basic statistics. Then, we illustrate the experimental setup, including the embedding size, initialization, etc. Afterward, all benchmark methods and evaluation metrics are introduced. Finally, we report the experimental results from four aspects: 1) evidence of topic effects and timeliness; 2) performance comparisons; 3) parameter sensitiveness; and 4) two case studies for attention visualizations.

### A. Dataset Preparation

We collected a dataset including 372 818 questions and 1 739 222 answers associated with topics, upvotes, timestamps, etc., from Quora using the approach described in [1]. For training a robust model, we only reserved the questions with more than ten answers among which the most number of upvotes is over 20. Then, we also removed questions and answers which had less than ten words or questions without topics. After that we removed the questions whose views and upvotes were not stable (the posting time of Q&A was less than one month before the collect-data time). After data cleaning, there are 9353 questions and 218 965 answers left. Table II shows the detailed statistics of the dataset after preprocessing. Finally, we create two kinds of ground truth for answer selection and answer ranking, respectively. For answer selection,

### TABLE II
### STATISTICS OF THE DATASET

| Statistics | Value |
|---|---|
| Total number of questions | 9,353 |
| Total number of answers | 218,965 |
| Average number of answers per question | 23.4 |
| Average number of words per question | 35.6 |
| Average number of words per answer | 108.3 |
| Average number of sentences per answer | 6.1 |
| Average number of topics per question | 4.7 |
| Average number of triples per question | 190.7 |

### TABLE III
### STATISTICS OF THE TRAINING AND TESTING DATASET

| Data split | | The ratio of questions in training sets | | | |
|---|---|---|---|---|---|
| | | 90% | 80% | 70% | 60% |
| Training | Question | 8,417 | 7,482 | 6,547 | 5,611 |
| | Answer | 197,068 | 175,857 | 155,183 | 130,493 |
| | Triples | 1,605,454 | 1,433,493 | 1,270,750 | 1,009,476 |
| Testing | Question | 936 | 1,871 | 2,806 | 3,742 |
| | Answer | 21,897 | 43,108 | 63,782 | 88,472 |

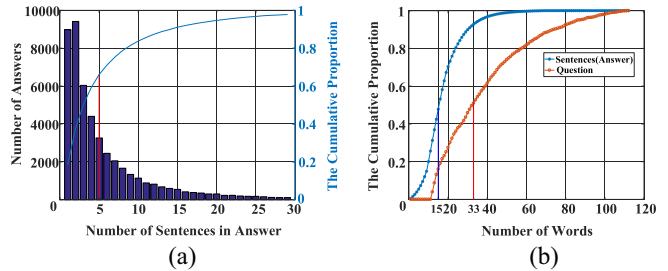

Fig. 4. Distribution of (a) sentences and (b) words.

we sorted answers to each question by their stable upvotes. Then, answers with more than ten upvotes were treated as good answers and the rest of them were treated as bad ones (i.e., ground truth of good answers equals to 1; 0 otherwise). For answer ranking, we sorted the answers of each question by their final upvotes and treated that ranking as the ground truth [1], [7].

Moreover, we also analyze some distributions of the number of sentences and words. Fig. 4 tells that about 40% answers contain more than 5 sentences and 50% questions (sentences in answers) contain more than 33 (15) words. That is to say, it is exhausting for the asker and visitors to read dozens of long answers to one question, so it is necessary to rank answers by their attraction to readers. In the following experiments, to observe how the models behave at different sparsity data, we randomly select 90%, 80%, 70%, and 60% instances as training sets and the rest as testing sets, respectively. More details are shown in Table III.

### B. Experimental Setup

*Embedding Setting:* Word embeddings in input layer are pretrained on the Q&A corpus in the whole collected data. We use public word2vec lib (Gensim[5]) to assign every word with a 50-dimensional vector (i.e., $K = 50$) which is tuned in training. Particularly, those words which appear less than five times are assigned a same randomly initialized vector. Besides, we empirically set the size of vectors $\boldsymbol{h}_t$, $z_t$ in (1) and $\boldsymbol{u}$ in (6) as 50.

---

[5]http://radimrehurek.com/gensim/

*Training Setting:* We follow [54] and randomly initialize all parameters in *EARNN* with a uniform distribution in the range from $-\sqrt{6/(\text{nin} + \text{nout})}$ to $\sqrt{6/(\text{nin} + \text{nout})}$, where nin and nout are the sizes of layers before and after the weight matrix. During the training process, all parameters are tuned. Moreover, we use dropout with probability 0.2 to prevent overfitting. Without special illustration, $H$ in (7) and $m$ in (8) are empirically set as $10^6$ (with unit second) and 0.1, respectively.

*Benchmark Methods:* In the experiments, *EARNN* represents our complete solution with word-level attention and also the time-sensitive ranking function. In order to illustrate the effectiveness of sentence-level and word-level attention mechanisms and the effects of timeliness, we construct two variant models, denoted by *EARNN_s* and *EARNN_w*. Specifically, *EARNN_s* only utilizes the sentence-level attention whose inputs are the details of Q&A. *EARNN_w* only utilizes the word-level attention which treats extra topics as an additional input. However, both variants treat the deep semantic matching score $\hat{V}$ as the ranking score $\widetilde{V}$ of the answer without considering the timeliness of answers.

Besides, we compare our approaches against six popular models for two tasks.

1) *BM25* [55] is a popular model in information retrieval. Correspondingly, the text of question can be treated as the query in *BM25* and answers are the documents to be ranked.
2) *TRLM* [29] is a translation-based model to calculate the similarity between two texts (i.e., the details of the question and answers).
3) *rankSVM* [56] is an SVM for ranking. Each question and answer is presented by a vector where each dimension denotes a word and its value equals to the frequency.
4) *NBOW* is a neural bag-of-words model where Q&A embeddings are the syntactic-level representations and a multilayer perceptron (MLP) [57] is used to measure the relevance.
5) *CNTN* [37] is a convolutional neural tensor network architecture which encodes Q&A and models their relevance with a tensor layer.
6) *WEC_CNN* [6] is the similarity matrix-based architecture to model the deep interactions between Q&A.

Among them, *BM25*, *TRLM*, and *RankSVM* are traditional methods using discrete word representations. The other deep learning-based methods use distributed vectors to model sentences. Particularly, *NBOW* and *CNTN* use syntactic- and semantic-level representations, respectively. And *WEC_CNN* is one of the deep interaction methods to match Q&A.

In the experiments, all methods are implemented by ourselves following the related references and all hyperparameters are tuned carefully so that their performances reach the best in Quora dataset. All results are obtained on a Linux System (4 Intel Core i5-6500 CPUs, 8 GB RAM). We use Tensorflow to implement deep learning methods (*NBOW*, *CNTN*, *WEC_CNN*, *EARNN_s*, *EARNN_w*, and *EARNN*).

Except the nonparameter models (i.e., *BM25* and *TRLM*), we also test the training time for *EARNN*, *WEC_CNN*, *CNTN*, *NBOW*, and *rankSVM* with the 90%–10% partitioning data. It, respectively, takes 7986 s, 960 s, 1242 s, 224 s, and 321 s for them to converge.

*Evaluation Metrics:* For evaluation, we rank answers as their predicted scores and compare the rank list with the ground truth (0/1 for answer selection task; ranking as their real upvotes for answer ranking task) to compute metrics.

*Answer Selection:* Since each answer is only noted by "good" or "bad" (i.e., 1 or 0), we adopt three types of metrics widely used in information retrieval. They are precision at $K$ ($P@k$), mean average precision (*MAP*), and mean reciprocal rank (*MRR*). Specifically, given the rank list, $P@k$ measures the precision on the top-$k$ answers. *MAP* is the mean of the average precision scores and *MRR* is the position of the first good answer in the candidate list. Formally, for each question, $P@k$, *MAP*, and *MRR* are defined as

$$P@k = \frac{1}{k} \sum_{j=1}^{k} I(j)$$

$$MAP = \frac{1}{|Q|} \sum_{j=1}^{N} I(j) \frac{\sum_{r=1}^{j} I(r)}{j}$$

$$MRR = \frac{1}{\text{rank}(j)} \tag{9}$$

where $I(j)$ is a binary function on the $j$th answer in the rank list and it equals 1 when the $j$th answer is good, 0 otherwise; $|Q|$ is the number of good answers, $N$ is the number of all answers, and rank($j$) is the position of the first good answer in the rank list. These three metrics range from 0 to 1 and the larger the better. In this paper, we choose $P@5$, $P@10$, *MAP*, and *MRR* for evaluation.

*Answer Ranking:* For this task, we adopt two widely used ranking metrics. One names normalized discount cumulative gain (*NDCG@k*) [58], where $k$ represents the top-$k$ ranked answers. Formally, for each question, *NDCG@k* is defined as follows:

$$NDCG@k = \frac{DCG@k}{iDCG@k} \tag{10}$$

where $iDCG$ is the ideal $DCG$ and $DCG$ is defined as

$$DCG@k = \text{rel}_1 + \sum_{j=2}^{k} \frac{\text{rel}_j}{\log_2 j} \tag{11}$$

where $\text{rel}_j$ equals to the rating of the $j$th answer.

Considering that *NDCG@k* measures the ranking quality of top-$k$ answers, we use another metric named degree of agreement (*DOA*) [59] which can measure the quality of an entire ranking list. Specifically, for a list of answers to a question, we assume $(x_i, y_i)$ is the observation and evaluation rank of the $i$th answer. Any pair of $(x_i, y_i)$ and $(x_j, y_j)$, where $y_i - y_j > 0$, is said to be a correct order pair if $x_i - x_j \geq 0$. Then, for each question, *DOA* is defined as

$$DOA = \frac{2n_c}{n(n-1)} \tag{12}$$

TABLE IV
Performances of Answer Selection on Four Metrics. Results Are Divided Into Four Parts According to the Ratio of Questions in Training Sets. Upper Left: 90%, Upper Right: 80%, Bottom Left: 70%, and Bottom Right: 60%

| | P@5 | P@10 | MAP | MRR | | P@5 | P@10 | MAP | MRR |
|---|---|---|---|---|---|---|---|---|---|
| *BM25* | 0.287 | 0.252 | 0.420 | 0.529 | *BM25* | 0.283 | 0.245 | 0.417 | 0.527 |
| *TRLM* | 0.277 | 0.223 | 0.428 | 0.582 | *TRLM* | 0.278 | 0.229 | 0.419 | 0.576 |
| *rankSVM* | 0.338 | 0.267 | 0.505 | 0.662 | *rankSVM* | 0.332 | 0.264 | 0.498 | 0.648 |
| *NBOW* | 0.338 | 0.266 | 0.495 | 0.640 | *NBOW* | 0.336 | 0.270 | 0.505 | 0.638 |
| *CNTN* | 0.350 | 0.270 | 0.513 | 0.650 | *CNTN* | 0.344 | 0.270 | 0.509 | 0.647 |
| *WEC_CNN* | 0.355 | 0.273 | 0.514 | 0.656 | *WEC_CNN* | 0.352 | 0.272 | 0.514 | 0.653 |
| *EARNN_s* | 0.361 | 0.274 | 0.532 | 0.673 | *EARNN_s* | 0.354 | 0.273 | 0.525 | 0.669 |
| *EARNN_w* | 0.365 | 0.275 | 0.539 | 0.688 | *EARNN_w* | 0.352 | 0.275 | 0.533 | 0.677 |
| ***EARNN*** | **0.404** | **0.303** | **0.587** | **0.739** | ***EARNN*** | **0.390** | **0.295** | **0.573** | **0.724** |
| | P@5 | P@10 | MAP | MRR | | P@5 | P@10 | MAP | MRR |
| *BM25* | 0.285 | 0.247 | 0.422 | 0.527 | *BM25* | 0.281 | 0.246 | 0.420 | 0.526 |
| *TRLM* | 0.270 | 0.224 | 0.417 | 0.570 | *TRLM* | 0.273 | 0.222 | 0.414 | 0.567 |
| *rankSVM* | 0.330 | 0.262 | 0.495 | 0.645 | *rankSVM* | 0.339 | 0.272 | 0.502 | 0.647 |
| *NBOW* | 0.336 | 0.269 | 0.502 | 0.635 | *NBOW* | 0.347 | 0.276 | 0.498 | 0.631 |
| *CNTN* | 0.348 | 0.272 | 0.510 | 0.643 | *CNTN* | 0.350 | 0.272 | 0.504 | 0.634 |
| *WEC_CNN* | 0.350 | 0.274 | 0.508 | 0.646 | *WEC_CNN* | 0.356 | 0.281 | 0.513 | 0.653 |
| *EARNN_s* | 0.353 | 0.274 | 0.522 | 0.664 | *EARNN_s* | 0.356 | 0.280 | 0.515 | 0.655 |
| *EARNN_w* | 0.358 | 0.276 | 0.531 | 0.675 | *EARNN_w* | 0.363 | 0.283 | 0.527 | 0.665 |
| ***EARNN*** | **0.386** | **0.294** | **0.573** | **0.723** | ***EARNN*** | **0.393** | **0.303** | **0.569** | **0.720** |

where $n_c$ is the number of correct order pairs and $n$ is the number of candidate answers. Both *NDCG@k* and *DOA* range from 0 to 1 and the larger the better. In this paper, we choose *NDCG@1*, *NDCG@5*, *NDCG@10*, and *DOA* for evaluation.

### C. Experimental Results

*1) Analysis of Topic Effects and Timeliness:* First, we report some evidences from Quora to strength our motivations. Generally, in CQA websites, when posting a question, the asker can choose a topic for it and this topic can help readers quickly find their favorite questions. Compared with those CQA sites whose topics are predefined such as Yahoo Answers, Quora provides an open way where the asker can create the topics for a specific question as shown in Fig. 1 (blue rectangle). Therefore, the asker usually extracts some key words from the question as topics which contain some intent of the asker. In Fig. 5(a), we plot the number of new topics per month by blue bars and also the number of all topics by the red line.[6] According to our statistics, since the establishment of Quora, the number of topics linearly increase at an average rate of 1600 per week and it reaches 70 thousands over 41 weeks. With the rapid growth of the number of topics, there is a wealth of semantic information in them such as coarse ones (e.g., "Movies" and "Startups") and fine ones (e.g., "Who Are the Best Professors at X"). Through the analysis of these topics, we can easily find the emphasis and hidden intention in the question. Then, the question can be better understood and it is beneficial for measuring the quality of answers.

Besides the topic effects, we also analyze the timeliness in CQA. In CQA website, after answering a question, the answer will be seen by all readers and visitors. Intuitively, with more and more users seeing it as time goes on, more upvotes are received even if the answer is not perfect. Therefore, when comparing two similar answers to one question, the early posted answer will receive more upvotes. In order to better illustration, we collect all upvotes of answers in a specific period and put them into different buckets according to the number of weeks they are posted (i.e., from one week to five weeks). Since most answers only obtain few upvotes and they have a bad effect on the analysis, we reserve answers with more than ten upvotes. Then, we draw a box plot for each bucket as shown in Fig. 5(b). For each box plot, top bar is maximum observation, lower bar is minimum observation, top of box is the third quartile, bottom of box is the first quartile, middle bar is median value, and red crosses are possible outliers. We can find the longer answers are posted, the larger the maximum upvotes. Similar phenomena occur on the third quartile and median value which indicate the analysis of timeliness is correct and worthy.

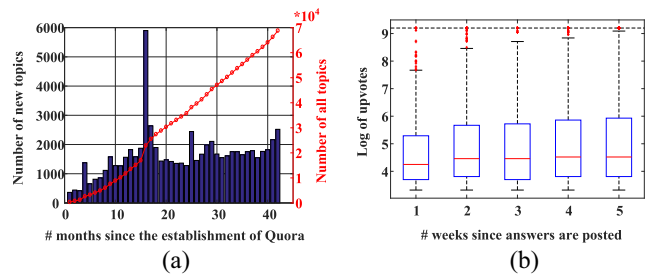

(a)                                    (b)

Fig. 5.  Evidences of our motivations. (a) Topic effects. (b) Timeliness.

*2) Performance Comparisons:* Second, we show the performance comparisons among all models on answer selection and answer ranking. For the former task, we list the results on *P@5*, *P@10*, *MAP*, and *MRR* in Table IV, while for the latter task, we list the results on *NDCG@1*, *NDCG@5*, *NDCG@10*, and *DOA* in Table V. As indicated in two

---

[6]Based on the statistics from https://neilpatel.com/blog/quora/, Quora experienced an estimated 150% growth of the number of unique visitors between December 2010 and January 2011 [i.e., 16th month in Fig. 5(a)]. Accordingly, the number of topics also increased rapidly around that month.

TABLE V
Performances of Answer Ranking on Four Metrics. Results Are Divided Into Four Parts According to the Ratio of Questions in Training Sets. Upper Left: 90%, Upper Right: 80%, Bottom Left: 70%, and Bottom Right: 60%

|  | NDCG@1 | NDCG@5 | NDCG@10 | DOA |  | NDCG@1 | NDCG@5 | NDCG@10 | DOA |
|---|---|---|---|---|---|---|---|---|---|
| BM25 | 0.556 | 0.623 | 0.711 | 0.560 | BM25 | 0.545 | 0.621 | 0.712 | 0.560 |
| TRLM | 0.611 | 0.629 | 0.700 | 0.523 | TRLM | 0.601 | 0.621 | 0.697 | 0.512 |
| rankSVM | 0.675 | 0.684 | 0.763 | 0.619 | rankSVM | 0.661 | 0.684 | 0.755 | 0.608 |
| NBOW | 0.644 | 0.686 | 0.756 | 0.623 | NBOW | 0.639 | 0.685 | 0.756 | 0.618 |
| CNTN | 0.650 | 0.695 | 0.763 | 0.625 | CNTN | 0.652 | 0.693 | 0.763 | 0.624 |
| WEC_CNN | 0.652 | 0.693 | 0.762 | 0.628 | WEC_CNN | 0.650 | 0.691 | 0.761 | 0.626 |
| EARNN_s | 0.677 | 0.707 | 0.772 | 0.636 | EARNN_s | 0.670 | 0.700 | 0.768 | 0.629 |
| EARNN_w | 0.680 | 0.711 | 0.775 | 0.640 | EARNN_w | 0.676 | 0.706 | 0.770 | 0.631 |
| **EARNN** | **0.720** | **0.731** | **0.793** | **0.651** | **EARNN** | **0.705** | **0.729** | **0.792** | **0.651** |
|  | NDCG@1 | NDCG@5 | NDCG@10 | DOA |  | NDCG@1 | NDCG@5 | NDCG@10 | DOA |
| BM25 | 0.544 | 0.620 | 0.710 | 0.557 | BM25 | 0.543 | 0.617 | 0.709 | 0.558 |
| TRLM | 0.603 | 0.623 | 0.698 | 0.514 | TRLM | 0.606 | 0.623 | 0.696 | 0.514 |
| rankSVM | 0.657 | 0.682 | 0.753 | 0.601 | rankSVM | 0.659 | 0.681 | 0.752 | 0.603 |
| NBOW | 0.629 | 0.683 | 0.755 | 0.614 | NBOW | 0.628 | 0.680 | 0.752 | 0.607 |
| CNTN | 0.648 | 0.695 | 0.766 | 0.625 | CNTN | 0.647 | 0.694 | 0.765 | 0.624 |
| WEC_CNN | 0.631 | 0.688 | 0.761 | 0.623 | WEC_CNN | 0.629 | 0.688 | 0.760 | 0.619 |
| EARNN_s | 0.652 | 0.698 | 0.766 | 0.630 | EARNN_s | 0.660 | 0.700 | 0.769 | 0.629 |
| EARNN_w | 0.666 | 0.704 | 0.772 | 0.632 | EARNN_w | 0.664 | 0.703 | 0.771 | 0.631 |
| **EARNN** | **0.695** | **0.726** | **0.791** | **0.648** | **EARNN** | **0.687** | **0.721** | **0.788** | **0.645** |

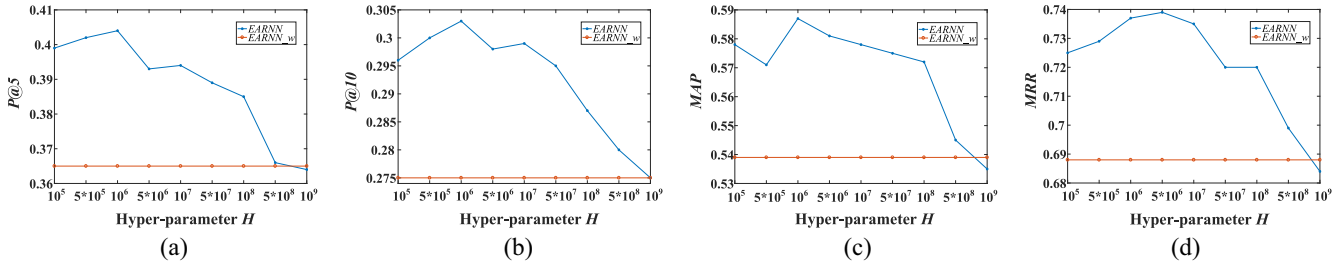

Fig. 6. Effects of hyperparameter $H$ on four metrics used in answer selection. (a) $P@5$. (b) $P@10$. (c) $MAP$. (d) $MRR$.

tables, since both tasks focus on measuring semantic matching between Q&A, the performances of them have shown the similar patterns. Our proposed models (i.e., *EARNN*, *EARNN_w*, and *EARNN_s*) outperform the baselines at most cases, indicating the effectiveness of our models on exploring topic effects and timeliness. Specific to our three models, *EARNN* performs best and *EARNN_w* ranks the second, followed by *EARNN_s*. Particularly, *EARNN_w* always performs better than *EARNN_s*, indicating the topic effects on answer ranking and our word-level attention mechanism can succeed in capturing the deep semantic relations between topics and Q&A. Except *EARNN*, all models mainly consider the similarities of Q&A and *EARNN_w* beat the best baseline (i.e., *WEC_CNN*) with the promotion of 1.8%, 0.8%, 4.0%, and 3.7% on $P@5$, $P@10$, $MAP$, and $MRR$ and 4.9%, 2.3%, 1.4%, and 1.5% on $NDCG@1$, $NDCG@5$, $NDCG@10$, and $DOA$. It suggests our models are better at measuring answer quality on our dataset. Besides, compared with *EARNN_w* and *EARNN*, respectively, increases 9.4%, 7.8%, 8.1%, and 7.4% on four metrics in answer selection and 4.5%, 2.9%, 2.5%, and 2.4% on four metrics in answer ranking so that we conclude that timeliness exists in CQA and our time-sensitive ranking function is effective on modeling this phenomenon.

Among the baselines, experimental results reveal the following points. First, in most cases, the performances of *NBOW*, *CNTN*, and *WEC_CNN* are better than those of *BM25* and *TRLM* indicating the powerful strength of deep learning models. However, as a conventional approach, *RankSVM* is still competitive with simple neural network methods, e.g., *NBOW*. Second, the performances of *NBOW* are not good enough compared with the performances of other neural network models, which demonstrates that CNN or RNN models can truly capture the semantic information in Q&A. Third, the observation that *WEC_CNN* performs quite well among the baselines shows that the interaction among words is effective to evaluate the relevance between sentences.

*3) Parameter Sensitiveness:* Here, we evaluate the sensitiveness of hyperparameter $H$ which can adjust the weight of the decay factor in (7). As mentioned in Section III-B, the decay factor is used to model the timeliness. When time goes on, the decay factor becomes smaller so that the value of the corresponding answer becomes lower and lower. Fig. 6 shows the performances on answer selection, and Fig. 7 shows the performances on answer ranking. Since we are intended to exploit the impact of the decay factor, we compare the performances between *EARNN* (blue curve) and *EARNN_w* (red curve) where there is only one difference on the prediction. In this part, we test performances with the 90%–10% partitioning data. Since the decay factor is close to 0 when $H$ is too small, we change hyperparameter $H$ beginning with $10^5$ to larger value. From the results, we notice that the performances of *EARNN* are better than those of *EARNN_w* in most cases. When $H$ varies from $10^5$ to $10^9$, the results appear a trend of going up first and then declining. Among different metrics, the performances achieve the peak at different point. However, the best $H$ value always ranges between $[10^6, 10^7]$. After the $H$ value exceeds $10^9$, the decay factor is close to 1 so that $EARNN$ degenerates into $EARNN\_w$.

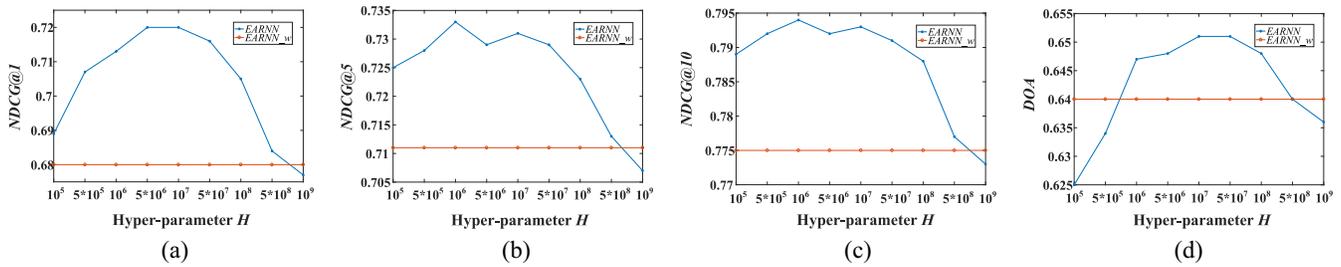

Fig. 7. Effects of hyperparameter $H$ on four metrics used in answer ranking. (a) $NDCG@1$. (b) $NDCG@5$. (c) $NDCG@10$. (d) $DOA$.

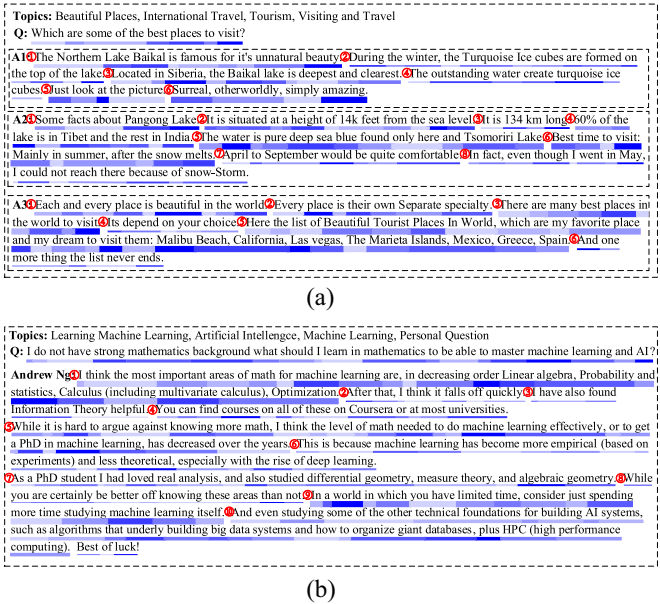

(a)

(b)

Fig. 8. Attention visualizations of the question and two answers. Each sentence begins with a number in red circle which denotes the current sentence. There is a stripe in blue beneath each word. The deeper the color, the more important the word is. The thicker the stripe, the more important the sentence is. (a) Visualization case of Q&A in Fig. 1. (b) Another visualization case answered by Andrew Ng.

*4) Case Study:* Here, we will illustrate one outstanding ability of *EARNN* on distinguishing the emphasis of Q&A, i.e., using attention scores generated by our word-level attention mechanism. Fig. 8 shows the sentence score $\alpha$ in (3) and the word score $\beta$ in (5) of Q&A with two cases. One is the motivating example in Fig. 1 and another is a suggestion from Andrew Ng for a fresher of machine learning.[7] Specifically, in order to clearly visualize, we classify all words (sentences) into several rates according to their word (sentence) scores $\alpha$ ($\beta$), i.e., the horizontal stripe beneath each word. The thickness of the stripe represents the importance of sentences and the depth of color represents the importance of words. Intuitively, the thicker the stripe, the deeper the color, the more important it is. In particular, the thickness of the stripe in questions makes no sense. From Fig. 8(a), we can easily find those words which can appeal to travelers are highlighted, such as beauty, "amazing," and many scenic spots. On the contrary, most prepositions and adverbs are assigned lower attention scores, such as "in," "on," "most," etc. Compared with the second answer **A2**, the first answer **A1** describes a surreal and amazing lake (i.e., the sixth sentence) and the third answer **A3** involves multiple scenic spots (i.e., the fifth sentence). These two answers are much better in terms of the content.

We also illustrate an answer from Andrew Ng for a fresher who is asking for help mastering machine learning and AI. From the machine understanding, it is convincing that the asker focuses on words like "mathematics," "machine learning," and "AI." In Andrew Ng's answer, machine finds the first and the ninth sentences are helpful to solve the question. Especially, the first sentence involves several mathematics, such as "linear algebra," "probability and statistics," "calculus," and "optimization." Although Andrew Ng refers to some opinions about machine learning (i.e., the fifth and the sixth sentences) and his own experience (i.e., the seventh sentence), machine does not treat them as important ones. These two visualizations illustrate that our two attention mechanisms can clearly capture the emphasis of Q&A which is beneficial for the development of CQA.

## V. SUMMARY AND FUTURE WORK

In this paper, we comprehensively inspected the answer selection and ranking problems by taking full advantage of both Q&A and multifacet domain effects. Particularly, we developed a serialized LSTM together with two enhanced attention mechanisms to model topic effects. Meanwhile, the emphasis of Q&A was automatically distinguished. We also designed a time-sensitive ranking function to establish the relations between Q&A and timeliness. We evaluated the performances of *EARNN* using the dataset from Quora and extensive experimental results clearly validated the effectiveness and interpretability of *EARNN*. In the future, we plan to generalize our model to those CQA systems whose topics are predefined such as Yahoo Answers. We would also like to exploit and model more domain effects based on our findings in CQA.

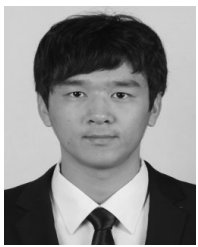

**Binbin Jin** received the B.E. degree in computer science from the University of Science and Technology of China, Hefei, China, in 2016, where he is currently pursuing the Ph.D. degree in computer science with the Department of Computer Science and Technology under the advisory of Prof. E. Chen.

He has published several papers in referred journals and conference proceedings, such as the IEEE TRANSACTIONS ON KNOWLEDGE AND DATA ENGINEERING and AAAI Conference on Artificial Intelligence. His current research interests include data mining, deep learning and applications in community question answering, and Internet finance-based websites (such as crowdfunding).

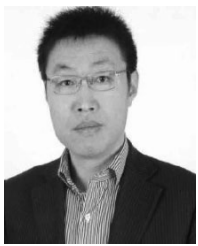

**Enhong Chen** (SM'07) received the Ph.D. degree in computer science from the University of Science and Technology of China (USTC), Hefei, China, in 1996.

He is a Professor and the Vice Dean of the School of Computer Science, USTC. His research is supported by the National Science Foundation for Distinguished Young Scholars of China. He has published over 100 papers in refereed conferences and journals, including the IEEE TRANSACTIONS ON KNOWLEDGE AND DATA ENGINEERING, IEEE TRANSACTIONS ON MOBILE COMPUTING, ACM SIGKDD Conference on Knowledge Discovery and Data Mining (KDD), IEEE International Conference on Data Mining (ICDM), Conference on Neural Information Processing Systems, and ACM International Conference on Information and Knowledge Management. His current research interests include data mining and machine learning, social network analysis, and recommender systems.

Dr. Chen was a recipient of the Best Application Paper Award on KDD-2008, the Best Research Paper Award on ICDM-2011, and the Best of SIAM International Conference on Data Mining (SDM)-2015. He was on the program committees of numerous conferences, including KDD, ICDM, and SDM.

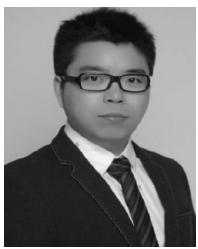

**Hongke Zhao** received the Ph.D. degree in computer science from the University of Science and Technology of China, Hefei, China, in 2018.

He is currently a Faculty Member of the College of Management and Economics, Tianjin University, Tianjin, China. He has published over 20 papers in refereed journals and conference proceedings, such as the *ACM Transactions on Intelligent Systems and Technology*, IEEE TRANSACTIONS ON SYSTEMS, MAN, AND CYBERNETICS: SYSTEMS, IEEE TRANSACTIONS ON BIG DATA, *Information Sciences*, ACM SIGKDD Conference on Knowledge Discovery and Data Mining (ACM SIGKDD), International Joint Conferences on Artificial Intelligence, AAAI Conference on Artificial Intelligence (AAAI), and IEEE International Conference on Data Mining. His current research interests include data mining, knowledge management, and Internet finance-based applications, such as crowdfunding and P2P lending.

Dr. Zhao was the Program Committee Member at ACM SIGKDD and AAAI.

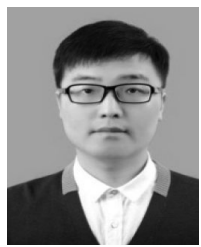

**Zhenya Huang** received the B.E. degree in software engineering from Shandong University, Jinan, China, in 2014. He is currently pursuing the Ph.D. degree in computer science with the School of Computer Science and Technology, University of Science and Technology of China, Hefei, China.

He has published several papers in referred conference proceedings, such as AAAI Conference on Artificial Intelligence, ACM International Conference on Information and Knowledge Management, International Conference on Database Systems for Advanced Applications, and SIGKDD. His current research interests include data mining and knowledge discovery, recommender systems, and intelligent education systems.

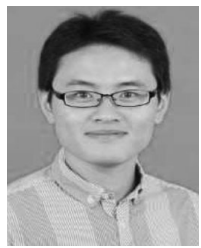

**Qi Liu** (M'15) received the Ph.D. degree in computer science from the University of Science and Technology of China (USTC), Hefei, China, in 2013.

He is an Associate Professor with USTC. He has published prolifically in refereed journals and conference proceedings, such as the IEEE TRANSACTIONS ON KNOWLEDGE AND DATA ENGINEERING, *ACM Transactions on Information Systems*, *ACM Transactions on Knowledge Discovery From Data*, ACM Transactions on Intelligent Systems and Technology, ACM SIGKDD Conference on Knowledge Discovery and Data Mining, International Joint Conferences on Artificial Intelligence, AAAI Conference on Artificial Intelligence, IEEE International Conference on Data Mining (ICDM), SIAM International Conference on Data Mining (SDM), and ACM International Conference on Information and Knowledge Management. His current research interests include data mining and knowledge discovery.

Dr. Liu was a recipient of the ICDM 2011 Best Research Paper Award and the Best of SDM 2015 Award. He has served regularly in the program committees of a number of conferences. He is a reviewer for the leading academic journals in his research areas. He is a member of ACM.

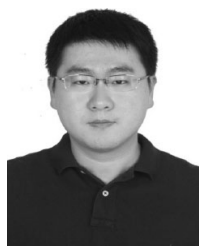

**Hengshu Zhu** (M'14) received the B.E. and Ph.D. degrees in computer science from the University of Science and Technology of China (USTC), Hefei, China, in 2009 and 2014, respectively.

He is currently a Senior Data Scientist with Baidu, Inc., Beijing, China. He has published prolifically in refereed journals and conference proceedings, including the IEEE TRANSACTIONS ON KNOWLEDGE AND DATA ENGINEERING, IEEE TRANSACTIONS ON MOBILE COMPUTING, *ACM Transactions on Knowledge Discovery From Data*, ACM SIGKDD Conference on Knowledge Discovery and Data Mining, International Joint Conferences on Artificial Intelligence, and AAAI Conference on Artificial Intelligence. His current research interests include data mining and machine learning, with a focus on developing advanced data analysis techniques for emerging applied business research.

Dr. Zhu was regularly on the program committees of numerous conferences. He has served as a reviewer for many top journals in relevant fields.

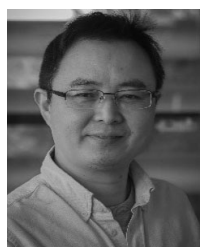

**Shui Yu** received the Ph.D. degree in computer science from Deakin University, Geelong, VIC, Australia, in 2004.

He is a Professor with the School of Computer Science, University of Technology Sydney, Sydney, NSW, Australia. He initiated the research field of networking for big data in 2013. His *H*-index is 37. He has published two monographs and edited two books, over 280 technical papers, including top journals and top conferences. His current research interests include security and privacy, networking, big data, and mathematical modeling.

Dr. Yu is currently serving on the editorial boards of the IEEE COMMUNICATIONS SURVEYS AND TUTORIALS (Area Editor), *IEEE Communications Magazine*, IEEE INTERNET OF THINGS JOURNAL, IEEE COMMUNICATIONS LETTERS, IEEE ACCESS, and IEEE TRANSACTIONS ON COMPUTATIONAL SOCIAL SYSTEMS. He is a member of AAAS and ACM, and a Distinguished Lecturer of IEEE Communication Society.